\documentclass[lettersize,journal]{IEEEtran}
\usepackage{amsmath,amsfonts}
\usepackage{algcompatible}
\usepackage{algorithm}

\usepackage{array}
\usepackage[caption=false,font=normalsize,labelfont=sf,textfont=sf]{subfig}
\usepackage{textcomp}
\usepackage{stfloats}
\usepackage{url}
\usepackage{verbatim}
\usepackage{graphicx}
\usepackage{xcolor}

\usepackage{multirow}
\usepackage{amssymb, mathtools}
\def\superscript#1#2#3#4#5{\prescript{\scalebox{.7}{$#1$}}{\scalebox{.7}{$#2$}}{#3}^{\scalebox{.7}{$#4$}}_{\scalebox{.7}{$#5$}}}
\newcommand{\Hquad}{\hspace{0.5em}} 

\def\BibTeX{{\rm B\kern-.05em{\sc i\kern-.025em b}\kern-.08em
    T\kern-.1667em\lower.7ex\hbox{E}\kern-.125emX}}
\usepackage{balance}
\begin{document}
\title{Tightly-Coupled, Speed-aided Monocular Visual-Inertial Localization in Topological Map}
\author{Chanuk Yang, Hayeon O, Kunsoo Huh

}

\maketitle

\begin{abstract}
This paper proposes a novel algorithm for vehicle speed-aided monocular visual-inertial localization using a topological map. The proposed system aims to address the limitations of existing methods that rely heavily on expensive sensors like GPS and LiDAR by leveraging relatively inexpensive camera-based pose estimation. The topological map is generated offline from LiDAR point clouds and includes depth images, intensity images, and corresponding camera poses. This map is then used for real-time localization through correspondence matching between current camera images and the stored topological images. The system employs an Iterated Error State Kalman Filter (IESKF) for optimized pose estimation, incorporating 
correspondence among images and vehicle speed measurements to enhance accuracy. Experimental results using both open dataset and our collected data in challenging scenario, such as tunnel, demonstrate the proposed algorithm's superior performance in topological map generation and localization tasks.
\end{abstract}

\begin{IEEEkeywords}
Visual Localization, Map Matching, IESKF, Tightly Coupled
\end{IEEEkeywords}

\section{Introduction}
\IEEEPARstart{H}{igh} precision localization is a crucial module for controlling autonomous vehicles. Many research efforts in autonomous driving utilize various sensors to develop highly accurate localization. A prominent sensor used in the localization module is GPS (Global Positioning System). The GPS can estimate the vehicle's pose regardless of its state and minor environmental changes. Additionally, lidar sensors can be utilized for pose estimation. Lidar, with its high accuracy in detecting points, can create a static map using these point clouds to estimate position. However, GPS prices vary significantly based on accuracy, and GPS sensors supporting Real Time Kinematics (RTK) tend to be expensive. Furthermore, lidar becomes more expensive as the number of points recognized per frame increases. Hence, research utilizing relatively inexpensive cameras for pose estimation is actively pursued.

Pose estimation using cameras is exemplified by visual SLAM (Simultaneous Localization And Mapping) research. Visual SLAM can estimate pose using differences in features or intensities in camera images. With the advancement of visual SLAM research, real-time position estimation and mapping can be conducted simultaneously \cite{geneva2020openvins},\cite{qin2018vins} . However, using visual SLAM alone requires further research to achieve global consistency. There exist some research to achieve global consistency through loop closure \cite{qin2018vins},\cite{campos2021orb} , but it does not always guarantee global consistency when obtaining poses in real-time situations.

To achieve real-time pose with global consistency, a method involves utilizing lidar point cloud maps and aligning them with points mapped by visual SLAM. An exemplary algorithm used for this purpose is ICP (Iterative Closest Point) \cite{sun2019scale},\cite{yabuuchi2021visual},\cite{zhang2023cross} or NDT (Normal Distribution Transform) \cite{zuo2019visual}. ICP finds optimized transformations that minimize the differences of points from matching point-to-point or point-to-plane feature. NDT also finds transformations from spatial distribution of points in a point cloud using normal distributions. However, the operation time of ICP may vary, leading to delays. Additionally, when structured points from visual SLAM have errors, ICP matching may not be accurate. NDT is the alternative approach in ICP, but grid resolution trade-off in performance and computational load is inevitable. 

Many research on visual localization using lidar prior maps utilizes ICP or NDT matching \cite{sun2019scale},\cite{yabuuchi2021visual},\cite{zhang2023cross},\cite{zuo2019visual} . Many studies directly utilize poses in loosely coupled method for visual localization. This inevitably leads to delays and may not ensure real-time performance when implemented in actual vehicles.

This paper proposes vehicle speed aided monocular visual-inertial localization. Unlike previous research, lidar point cloud maps are transformed into a topological map format according to poses for map matching. Lidar depth image and the corresponding camera image are stored for each pose. A filtering algorithm is proposed to estimate poses through correspondence matching between the map and the current camera image. In the localization process, the image and correspondence stored in the current image are matched with the image from the topological map. Then, based on the matched points, the pose through the IESKF (Iterated Error State Kalman Filter) is estimated. Unlike other visual localization algorithms, feature tracking algorithms are not used in this study. In addition, the pose is estimated using the image feature in a tightly coupled form.

The contributions of this paper are as follows:
\begin{itemize}{}{}
\item {Proposing a topological map for matching images with lidar point cloud maps to make cross modal correspondence matching directly.}
\item {Introducing IESKF (Iterated Error State Kalman Filter) based visual localization without feature tracking method which needs additional optimization process.}
\item {Estimating poses using a tightly coupled map matching based on residuals between points from lidar map and features from image to improve the localization performance.}

\end{itemize}

\section{Related Works}

\subsection{Filtering-based Localization and Map Matching}

Filtering-based localization algorithms primarily consist of EKF (Extended Kalman Filter), UKF (Unscented Kalman Filter), or PF (Particle Filter). Tesli'c et al. \cite{teslic2011ekf} utilize wheel encoders and 2D LiDAR for localization. The correction step is performed by minimizing the difference between the matched line segments from the local and global maps. Allotta et al. \cite{allotta2015comparison} and D’Alfonso et al. \cite{d2015mobile} compare the performance of EKF and UKF-based localization in AUVs or mobile robots. Dellaert et al. \cite{dellaert1999monte} first introduced particle filter-based localization, known as Monte Carlo localization, for mobile robotics. Building on Monte Carlo localization, Akai et al. \cite{akai20203d} introduced 3D Monte Carlo localization utilizing 3D LiDAR. They also developed a reliable Monte Carlo localization with quick relocalization and reliability estimation \cite{akai2023reliable}.

Map matching can also be used for localization in various ways. Xia et al. \cite{xia2023integrated} utilized NDT based map matching with a light point cloud map and current LiDAR frame for localization. Kim et al. \cite{kim2022road} proposed a localization algorithm by matching HD maps with line segmentations detected from camera modules. They performed geometry-based map matching to account for matching failures and demonstrated superior performance compared to ICP. Wang et al. \cite{wang2017map} extracted curbs from 3D LiDAR data, accumulated them, and performed ICP matching with a digital map for localization. Sobreira et al. \cite{sobreira2019map} compared the performance of three commonly used map matching algorithms: ICP, NDT, and PM (Perfect Match). Sarlin et al. \cite{sarlin2023orienternet} introduced Orienternet, which performs map matching using open street maps and camera images through a CNN (Convolutional Neural Network).

\subsection{Frontend: Learning based Image Feature matching}
 Feature matching aims to find precise feature correspondences between different images. As these correspondences explain the geometric relationship between two images, feature matching across sequential images allows us to understand the geometric changes (epipolar geometry) over time. Before the advent of deep learning, a process of image matching can be decomposed into feature detection, feature description, feature matching, and geometric transformation estimation, as mentioned in \cite{xu2024local}. While these methods could withstand various transformations under certain theoretical conditions, they were fundamentally limited by the prior assumptions imposed by researchers on their tasks. The subsequent emergence of deep learning has addressed these limitations while providing robustness under various conditions. These approaches can be classified according to the learning method as follows.

Weakly-supervised learning is used when training with a small amount of labeled data and a large amount of unlabeled data. It is suitable for handling large-scale datasets, including the generation of topology maps in this paper, and is cost-efficient as it does not require densely annotated labels like fully-supervised methods. Instead of labels, these methods use camera poses \cite{wang2020learning, tyszkiewicz2020disk} or rewards to calculate the stability and repeatability of detected keypoints as supervision \cite{truong2021learning}. SuperPoint \cite{detone2018superpoint}, which is based on self-supervised learning, does not require any annotations. This method can be more cost-efficient than weakly-supervised methods but has the limitation of only being able to find corner points. In fully-supervised learning methods, where all datasets are labeled, high performance is ensured due to strong adaptability to learning new information from acquired datasets in defined scenarios. Unlike unsupervised learning, which requires large-scale datasets and exhibits high plasticity, thereby being less prone to overfitting, fully-supervised learning offers a more controlled approach. For instance, GLU-Net \cite{truong2020glu} extracts local and global features of images using CNNs, aiding in recognizing specific patterns in small areas and understanding the context of the entire image. LoFTR \cite{sun2021loftr} and ASpanFormer \cite{chen2022aspanformer} rely on Transformers and their attention mechanisms. LoFTR \cite{sun2021loftr} uses a self-attention mechanism to model relationships between positions in input images, while ASpanFormer \cite{chen2022aspanformer} adopts a hierarchical attention structure considering local-global context through a specific span structure. However, due to the complexity of these attention mechanisms, they may incur high computational costs. 

Among those methods, Patch2pix \cite{zhou2021patch2pix} adopts a weakly-supervised method using epipolar geometry as supervision. Detector-based methods are not robust to challenging scenarios, such as extreme viewpoint changes and textureless areas. 
Unlike previous models, Patch2pix \cite{zhou2021patch2pix} is detector-free, allowing it to directly extract visual descriptors and find consistent keypoints in image pairs. So it has been adopted for the offline process of generating topology maps due to its ability to provide cross-modality correspondence between depth images and intensity images. LightGlue \cite{lindenberger2023lightglue} addresses the computation cost issue by modifying the attention mechanism of SuperGlue \cite{sarlin2020superglue} based on graph neural networks. This modification reduces the computation cost by separating similarity and matchability of features in the prediction step from the baseline, thereby reducing the repeated computation cost of row-wise and column-wise normalization. As a result, it achieves superior performance while ensuring high speed, surpassing the accuracy of dense matchers for distributing points on dense grids, making it suitable for the online process discussed in this paper.

\subsection{Backend: Filtering-based SLAM}

Prominent filtering-based SLAM methods include MSCKF (Multi-State Constrained Kalman Filter), or IEKF (Iterated Extended Kalman Filter). Mourikis et al. \cite{mourikis2007multi} first introduced MSCKF, creating a tightly coupled visual-inertial odometry system. Li et al. \cite{li2013high} analyzed the observability issues in MSCKF and proposed methods to ensure consistent observability, also estimating IMU-Camera calibration parameters. Further, Sun et al. \cite{sun2018robust} employed stereo cameras with MSCKF to estimate poses. Geneva et al. \cite{geneva2020openvins} released an open-source version of MSCKF, demonstrating successful pose estimation across various datasets. Lee et al. \cite{lee2023mins} proposed an MSCKF algorithm that integrates not only camera data but also GPS and wheel odometry.

IEKF optimizes the state update through iterative correction processes. Bloesch et al. \cite{bloesch2017iterated} introduced an IEKF-based visual-inertial odometry, performing state updates using photometric errors from corner points extracted from images. Qin et al. \cite{qin2020lins} proposed the use of IEKF in LiDAR SLAM, offering a faster estimation method compared to traditional optimization approaches. Xu et al. \cite{xu2021fast} suggested a method to expedite the acquisition of Kalman gain values in tightly coupled IEKF with LiDAR points. In a subsequent work \cite{xu2022fast}, Xu improved the speed and performance of the mapping process by using an ikd-tree data structure for faster mapping.

In filtering-based SLAM, camera-based methods commonly involve feature tracking, with MSCKF being the preferred approach due to its effective utilization of this capability. In contrast, LiDAR-based SLAM typically does not involve feature tracking and primarily relies on IEKF methods, as evidenced by the research trends.
    
\subsection{Visual Localization from LiDAR Point Cloud Map}

There are methods for visual localization that utilize LiDAR point cloud maps for map matching. One approach involves matching feature maps created through visual SLAM with LiDAR point cloud maps. Sun et al. \cite{sun2019scale} conducted map matching by aligning points from monocular visual odometry with LiDAR point clouds, estimating the scale in the odometry to achieve the matching. Zuo et al. \cite{zuo2019visual} matched points from MSCKF-based visual odometry with a LiDAR prior map using NDT and investigated the sensitivity to inaccuracies in the prior map. Yabuuchi et al. \cite{yabuuchi2021visual} attempted map matching using only low-cost and lightweight cameras, demonstrating robustness against lighting and seasonal changes using real-world datasets. Furthermore, they integrated a lane center-line vector map for map matching and showed accurate position estimation in long-term localization \cite{yabuuchi2022vmvg}. Zhang et al. \cite{zhang2023cross} employed semantic consistency for point-to-plane ICP and decoupled the operation strategy to estimate affine transformation. The loosely coupled method that matches feature maps from visual SLAM with LiDAR point clouds and uses the resulting pose relies heavily on the accuracy of visual odometry. Inaccurate feature maps can lead to significant localization errors.

Another method involves map matching between LiDAR prior maps and images. Wolcott \cite{wolcott2014visual} created a cost map using NMI (Normalized Mutual Information) from multiple depth maps extracted from camera images and LiDAR prior maps to attempt map matching. Kim et al. \cite{kim2018stereo} matched depth images from stereo cameras with range images from LiDAR prior maps for pose estimation. Another approach involves tightly coupled pose estimation through feature matching. Caselitz et al. \cite{caselitz2016monocular} performed map matching by matching voxel grids from LiDAR point cloud maps with reconstructed points from visual SLAM. Yu et al. \cite{yu2020monocular} matched 2D lines from images with 3D lines from LiDAR maps for localization. Zheng et al. \cite{zheng2023tightly} conducted pose estimation by tracking and matching lines.

In contrast, the algorithm proposed in this paper does not perform map matching using ICP with feature maps from visual odometry. Instead, it employs deep learning for image correspondence matching. Rather than using the LiDAR prior map directly, depth images and intensity images corresponding to the global pose are extracted to create and use a topological map. The image correspondence matching is conducted with the current image, and the depth information is used to reconstruct 3D points from the matched 2D points. The localization algorithm tightly couples the features extracted from deep learning method without applying feature tracking algorithms.

\section{System overview}
The overall schematic diagram of the proposed algorithm is depicted in Figure 1. The generation of the topological map is an offline process, creating a prior map. The generated topological map is then utilized in the online localization process for real-time map matching. The detailed explanation of each process is described in Section IV and Section V. The topological map ($\mathcal{T}$) is structured as follows:

\begin{equation}
    \begin{array}{c}
        \mathcal{T} = \Bigl\{ (\superscript{C^*}{}{\mathcal{N}}{}{t}) \hspace{0.5em} | \hspace{0.5em}  t=1,...,T \Bigl\}
    \end{array}
\end{equation}

\begin{align}
    \begin{split}
        \superscript{C^*}{}{\mathcal{N}}{}{t} = \Bigl\{ &(\superscript{C^*}{}{\bold{D}}{}{t}, \superscript{C^*}{C}{\bold{I}}{}{t}, \superscript{C^*}{G}{\bold{T}}{}{t}) \hspace{0.5em}  \\ &|  \superscript{C^*}{}{\bold{D}}{}{t} \in \mathbb{R}^{w*h}, \superscript{C^*}{C}{\bold{I}}{}{t} \in \mathbb{R}^{w*h},  \superscript{C^*}{G}{\bold{T}}{}{t} \in SE(3) \Bigl\}
    \end{split}
\end{align}
where $\superscript{C^*}{}{\bold{D}}{}{t}$, $\superscript{C^*}{C}{\bold{I}}{}{t}$, and $\superscript{C^*}{G}{\bold{T}}{}{t}$ represent the depth image, camera image, and transformation, respectively, and the three are grouped into a single node ($\superscript{C*}{}{\mathcal{N}}{}{t}$). At the given time step, the corresponding camera image and depth image can be found based on the transformation. Important notations for the system formulations are listed in Table \ref{table_notation}.
 
\begin{figure*}[!t]
\centering
\includegraphics[width=7.0in]{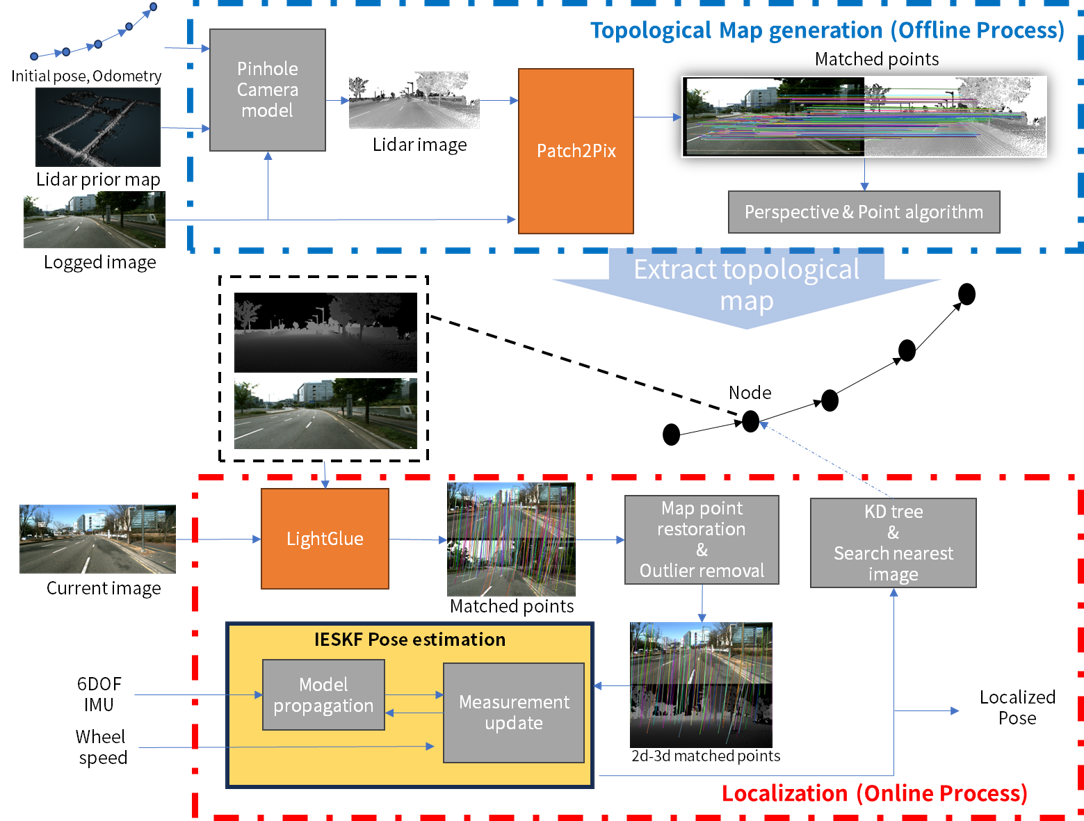}\label{fig_first_case}
\hfil
\caption{Overall schematic diagram of the proposed algorithm.}
\label{fig_sim}
\end{figure*}

\begin{table}
    \caption{Important notations for this paper}
    \label{table_notation}
    \centering
    \begin{tabular}{cc}
    \hline
        Symbols & Meaning\\
        \hline
        $C, C^*$ & Camera or camera frame, and groud truth camera frame \\
        $\mathcal{L}$ & Lidar prior map \\
        $G$ & Global frame \\
        $I$ & IMU frame \\
        $\superscript{A}{C}{\bold{I}}{}{t}$ & Type $C$(Camera) image in frame $A$ at timestep $t$\\
        $\superscript{A}{L}{\bold{I}}{}{t}$ & Type $L$(Lidar intensity) image in frame $A$ at timestep $t$\\
        $\superscript{A}{}{\bold{D}}{}{t}$ & Depth image in frame $A$ at timestep $t$ \\
        $\superscript{A}{B}{\bold{T}}{}{t}$ & Transformation from frame $A$ to frame $B$ at timestep $t$ \\
        $\superscript{A}{B}{\bold{R}}{}{t}$ & Rotation of frame $A$ in frame $B$ at timestep $t$. \\
        $\superscript{A}{B}{\bold{p}}{}{t}$ & translation of frame $A$ in frame $B$ at timestep $t$. \\
        $\superscript{A}{}{\bold{f}}{}{i}$ & $i$-th image feature location in frame $A$. \\
        $\superscript{A}{}{\bold{m}}{}{i}$ & $i$-th 3d map point in frame $A$.  \\
     \hline

    \end{tabular}
\end{table}

\section{Topological map generation}
In this study, LiDAR point cloud prior map is transformed into a topological map for efficient utilization. The algorithm for generating the topological map is presented in Algorithm 1. To convert into a topological map, the following inputs are required: camera image ($\superscript{C^*}{C}{\bold{I}}{}{k}$), camera intrinsic parameters ($\bold{K}$), point cloud map ($\mathcal{L}$), and global initial transformation ($ \superscript{G}{C}{\bold{T}}{}{k}$). In the first step of Algorithm 1, the initial transformation is used to rasterize the prior map, and the rasterized map is then projected using the pinhole camera model with the intrinsic parameters ($\bold{K}$) to match with the view of the stored camera image. This process generates the point cloud intensity image ($\superscript{C}{L}{\bold{I}}{}{k}$) and the depth image ($\superscript{C}{}{\bold{D}}{}{k}$).

In the second step of Algorithm 1, the Patch2Pix algorithm \cite{zhou2021patch2pix} is used to find the matching features ($\mathcal{F}_{L}$) in the point cloud intensity image and the corresponding features ($\mathcal{F}_{C*}$) in the camera image. The camera image and the projected point cloud image are utilized to find the camera transformation. This requires finding the correspondence matching points between the two images, a challenging task for multi-modal based correspondence matching algorithms. Patch2Pix is a deep learning-based correspondence matching algorithm that demonstrates the feasibility of multi-modal correspondence matching.

In the third step, the 2D matching points found in the point cloud intensity image are used to find the corresponding 3D points ($\mathcal{P}_{L}$) through the depth image. In the fourth step, outliers in the correspondence matching are removed using the estimate rotation RANSAC algorithm \cite{koide2023general}. The remaining 2D features ($\mathcal{F}'_{C*}$) and 3D points ($\mathcal{P}'_{L}$) are used in the Perspective and Point (PnP) algorithm \cite{terzakis2020consistently} to determine the transformation ($\superscript{C}{C^*}{\bold{T}}{}{}$) from the predicted camera transformation to the optimal camera transformation. This allows the calculation of the transformation between the initial global transformation and the camera transformation. The perspective and point algorithm used in this study is SQPnP \cite{terzakis2020consistently} from the OpenCV library \cite{opencv_library}. The calculated transformation is then multiplied by the initial global transformation to obtain the camera transformation, as shown in the following equation:
\begin{equation}
\superscript{C^*}{G}{\bold{T}}{}{t} = \superscript{C}{G}{\bold{T}}{}{t} \superscript{C}{C^*}{\bold{T}}{-1}{}   \label{tf_camera}
\end{equation}
where $\superscript{C^*}{G}{\bold{T}}{}{t}$, $\superscript{C}{G}{\bold{T}}{}{t}$, and $\superscript{C}{C^*}{\bold{T}}{-1}{}$ are the optimal camera transformation, the predicted camera transformation, and inverse of the transformation calculated from PnP algorithm, respectively. The obtained camera transformation is stored as a node in the topological map, along with the corresponding camera image. Additionally, a new depth image based on the camera transformation is calculated and stored in the previously created node. The position of the transformation within the node is stored in a kd-tree.

To create a topological map, a good initial transformation is required. However, it is challenging to ensure a good initial transformation for each node in the topological map. To address this, we use odometry to determine the initial transformations for sequentially arranged nodes. The camera transformation determined from the initial transformation can be calculated using the odometry difference with the next sequential node as described below: 
  
\begin{equation}
\superscript{C}{G}{\bold{T}}{}{k+1} = \superscript{C^*}{G}{\bold{T}}{}{k} (\superscript{C}{B}{\bold{T}}{-1}{}) 
                                (\superscript{B}{B_0}{\bold{T}}{-1}{k}\superscript{B}{B_0}{\bold{T}}{}{k+1}) \superscript{C}{B}{\bold{T}}{}{} \label{tf_next_pose} \\[7pt]
\end{equation}
where $\superscript{B}{B0}{\bold{T}}{}{k}$ and $\superscript{C}{B}{\bold{T}}{}{k}$ are transformation of the odometry and the extrinsic parameter from the camera to baseline, respectively. This approach enables the creation of the topological map using the initial transformation and odometry.

\begin{algorithm}[H]
\caption{Topological map generation}\label{alg_map_gen}
\begin{algorithmic}[1]
    \Statex \textbf{Input:} LiDAR point cloud prior map $\mathcal{L}$;
    \Statex \quad \quad \quad Camera image $\superscript{C^*}{C}{\bold{I}}{}{i}$;
    \Statex \quad \quad \quad Camera intrinsic parameter $\bold{K}$;
    \Statex \quad \quad \quad Predicted camera pose $\superscript{G}{C}{\bold{T}}{}{k}$
    \STATE $\superscript{C}{L}{\bold{I}}{}{k}, \superscript{C}{}{\bold{D}}{}{k} \gets$ get intensity, depth image from $\mathcal{L}, \bold{K}, \superscript{G}{C}{\bold{T}}{}{k}$
    \STATE$ \superscript{}{}{\mathcal{F}}{}{L},  \superscript{}{}{\mathcal{F}}{}{C*} \gets$ correspondence matching from $ \superscript{C}{L}{\bold{I}}{}{k},  \superscript{C^*}{C}{\bold{I}}{}{k}$
    \STATE$ \superscript{}{}{\mathcal{P}}{}{L} \gets$ map points extract from $ \superscript{}{}{\mathcal{F}}{}{L}, \superscript{C}{}{\bold{D}}{}{k} $
    \STATE$ \superscript{}{}{\mathcal{P}'}{}{L}, \superscript{}{}{\mathcal{F}'}{}{C*} \gets$ outlier removal in estimate rotation RANSAC
    \STATE$ \superscript{C^*}{C}{\bold{T}}{}{} \gets $ PnP solution from $\superscript{}{}{\mathcal{P}'}{}{L}, \superscript{}{}{\mathcal{F}'}{}{C*}$
    \STATE$ \superscript{G}{C^*}{\bold{T}}{}{k} \gets $ calculate global transformation via \eqref{tf_camera}
    \STATE$ \superscript{C^*}{}{\bold{D}}{}{k} \gets$ get depth image from $\mathcal{L}, \bold{K}, \superscript{G}{C^*}{\bold{T}}{}{k}$
    \Statex \textbf{Output:}  $\superscript{C^*}{}{\mathcal{N}}{}{k} = (\superscript{C^*}{}{\bold{D}}{}{k}, \superscript{C^*}{C}{\bold{I}}{}{k}, \superscript{C^*}{G}{\bold{T}}{}{k})$
\end{algorithmic}
\end{algorithm}

\section{Localization process}
After constructing the topological map, the localization process uses the map for map matching. In this study, tightly coupled map matching is performed using the correspondence matching between the current image and the images stored in the topological map. The matching points obtained from the correspondence matching algorithm are used as measurement data for the iterated Kalman filter.

\subsection{Correspondence Matching with outlier removal}
Assume that the current pose or initial pose is known. The pose can be used to find adjacent camera poses and camera images stored in the topological map's kd-tree. This allows for matching the current camera image with the camera image from the topological map. In this study, a deep learning-based correspondence matching algorithm, specifically LightGlue \cite{lindenberger2023lightglue}, is used for fast computation of the matched feature points. $\superscript{C}{}{\bold{f}}{}{i}$ is the matched feature point in the current image, and $\superscript{C*}{}{\bold{f}}{}{i}$ is the detected feature in the topological map image. The matching sets can be expressed as follows:

\begin{equation}
\mathcal{F}_C = \{ \superscript{C}{}{\bold{f}}{}{i} \hspace{0.5em} | \hspace{0.5em} \superscript{C}{}{\bold{f}}{}{i} \in \mathbb{R}^2, i=1, ..., N,N+1, ..., N+M  \}
\end{equation}
\begin{equation}
\mathcal{F}_{C*} = \{ \superscript{C*}{}{\bold{f}}{}{i} \hspace{0.5em} | \hspace{0.5em} \superscript{C*}{}{\bold{f}}{}{i} \in \mathbb{R}^2, i=1, ..., N,N+1, ..., N+M  \}
\end{equation}
where $N$ is the number of correctly matched points, and $M$ is the number of outliers.

Even after the correspondence matching, some points may be incorrectly matched. To prevent this, a rotation-only estimation-based RANSAC method \cite{koide2023general} can be used for more accurate outlier removal. However, this method is computationally intensive, especially with a high number of feature points and iterations, implying a trade-off between outlier accuracy and computation speed. 

In this study, a statistical method is adopted for outlier removal, considering computational efficiency. When the vehicle moves linearly, closer objects in the image have faster-moving feature points, while farther objects have slower-moving feature points. In the case of rotation, feature points move uniformly regardless of the object's distance. Assuming that the vehicle's motion is similar to the average motion of feature points, those points deviating beyond a defined variance are considered as outliers. The transformation of $\superscript{C*}{}{\bold{f}}{}{i}$ to the current image requires projection, expressed as follows:

\begin{equation}
\mathcal{F}_{C*}' = \Bigl\{ \superscript{C*}{}{\bold{f}}{'}{i} =  \pi_1(\pi^{-1}_2(\superscript{C*}{}{\bold{f}}{}{i})) \hspace{0.5em} \bigl| \hspace{0.5em} \forall \superscript{C*}{}{\bold{f}}{}{i} \in \mathcal{F}_{C*} \Bigl\}
\label{eq_F_C_start_dash}
\end{equation}
where$\superscript{C*}{}{\bold{f}}{'}{i}$ and $\pi^{-1}_2()$ are reprojected feature from the image of the topological map to the current image and the unprojection of the feature to a point, respectively. The unprojection utilizes the depth map stored with the topological map. $\pi_1()$ is the projection function to the current image, derived from the camera intrinsic parameters.

If $\superscript{C*}{}{\bold{f}}{'}{i} = [\superscript{C*}{}{u}{'}{i}, \superscript{C*}{}{v}{'}{i}]$ and $\superscript{C}{}{\bold{f}}{}{i} = [\superscript{C}{}{u}{}{i}, \superscript{C}{}{v}{}{i}]$, the outlier removal is expressed as follows:

\begin{equation}
\begin{array}{l}

u_m = \sum_{i=1}^{N+M} (\superscript{C*}{}{u}{'}{i}- \superscript{C}{}{u}{}{i}) / (N+M) \\  \\
v_m = \sum_{i=1}^{N+M} (\superscript{C*}{}{v}{'}{i}- \superscript{C}{}{v}{}{i}) / (N+M)
\end{array}
\label{eq_mean_pixel}
\end{equation}

\begin{equation}
    \begin{array}{c}
        \mathcal{S} = \Bigl\{ (\superscript{C*}{}{\bold{f}}{'}{i}, \superscript{C}{}{\bold{f}}{}{i} ) \hspace{0.5em} \Bigl| \hspace{0.5em}
        \superscript{C*}{}{\bold{f}}{'}{i} \in \mathcal{F}_{C*}',  \superscript{C}{}{\bold{f}}{}{i} \in \mathcal{F}_{C}, \\
        |(\superscript{C*}{}{u}{'}{i} - \superscript{C}{}{u}{}{i}) - u_m |< 3\sigma_{th}, \\ 
        |(\superscript{C*}{}{v}{'}{i} - \superscript{C}{}{v}{}{i}) - v_m| < 3\sigma_{th}  \Bigl\}
    \end{array}
\label{eq_matching_set_2d_2d}
\end{equation}
where $\superscript{C*}{}{u}{'}{i}$, $\superscript{C*}{}{v}{'}{i}$, $\superscript{C}{}{u}{}{i}$, and $ \superscript{C}{}{v}{}{i}$ are feature locations in image plane. $u_m$ and $v_m$ are mean of feature location difference between the reprojected features from topological map and the features from current image. The $S$ is inlier set of the correspondence matching and $n(S) = N $. The $\sigma_{th}$ is the user-defined threshold. This equation uses a normal distribution, treating points whose distance differences exceed three times the defined variance as outliers. To use the camera measurement data, the map data in global 3D-point values is expressed as follows:

\begin{equation}
    \superscript{G}{}{\bold{m}}{}{i} =  \superscript{C^*}{G}{\bold{R}}{}{} (\pi_2^{-1}(\superscript{C*}{}{\bold{f}}{'}{i})) + \superscript{C^*}{G}{\bold{p}}{}{}
\end{equation}
 
\begin{equation}
    \mathcal{S}' = \Bigl\{ (\superscript{G}{}{\bold{m}}{}{i},  \superscript{C}{}{\bold{f}}{}{i}) \hspace{0.5em} \bigl| \hspace{0.5em} \forall (\superscript{C*}{}{\bold{f}}{'}{i}, \superscript{C}{}{\bold{f}}{}{i}) \in \mathcal{S}  \Bigl\}
\end{equation}
where $\superscript{G}{}{\bold{m}}{}{i}$ and $\mathcal{S}'$ are 3d map point and set of 3d-2d correspondence matching, respectively. The $\superscript{C*}{G}{\bold{R}}{}{}$ and $\superscript{C*}{G}{\bold{p}}{}{}$ are the rotation and translation extracted from $\superscript{C*}{G}{\bold{T}}{}{k}$. 
\subsection{Model description}

\subsubsection{IMU kinematic model}
The discrete Inertial Measurement Unit (IMU) model for the nominal state can be expressed based on accelerometer data, $\bold{a}_m$, and gyroscope data, $\bold{w}_m$, \cite{sola2017quaternion}.

\begin{equation}
\renewcommand{\arraystretch}{1.2} \begin{array}{l}
        \superscript{I}{G}{\bar{\bold{R}}}{}{t+1} =  \superscript{I}{G}{\bold{R}}{}{t} \exp ([(\bold{w}_m-\bold{b}_{w,t})\Delta t]_\times) \\
        \superscript{I}{G}{\bar{\bold{p}}}{}{t+1} = \superscript{I}{G}{\bold{p}}{}{t} +\superscript{I}{G}{\bold{v}}{}{t}\Delta t + 0.5(\superscript{I}{G}{\bold{R}}{}{t}(\bold{a}_m-\bold{b}_{a,t}))\Delta t^2 \\
        \superscript{I}{G}{\bar{\bold{v}}}{}{t+1} = \superscript{I}{G}{\bold{v}}{}{t} + (\superscript{I}{G}{\bold{R}}{}{t}(\bold{a}_m-\bold{b}_{a,t}))\Delta t \\
        \Bar{\bold{b}}_{a,t+1} = \bold{b}_{a,t} \\
        \Bar{\bold{b}}_{w,t+1} = \bold{b}_{w,t} \\
        \Bar{\bold{g}}_{t+1} = \bold{g}_{t} \\
    
    \end{array}
\label{eq_imunominal}\end{equation}
where $\superscript{I}{G}{\bold{R}}{}{t}$ is the rotation matrix, $\superscript{I}{G}{\bold{p}}{}{t}$ is the position, $\superscript{I}{G}{\bold{v}}{}{t}$ is the velocity. $\bold{b}_{a,t}$, $\bold{b}_{w,t}$ and $\Delta t$ are the biases for the accelerometer, gyroscope at timestep $t$ and difference of timestep, respectively. $[\cdot]_\times$ is skew symmetric matrix operator of vectors. The terms with $\Bar{\cdot}$  (bar) means the predicted variables from the nominal IMU model. The discrete error state model is shown below from the continuous-time model \cite{sola2017quaternion}:

\begin{equation}
\renewcommand{\arraystretch}{1.2} \begin{array}{l}
        \superscript{I}{G}{\delta \boldsymbol{\theta}}{}{t+1} =  (\exp[(\bold{w}_m-\bold{b}_{w,t})\Delta t]_\times)^T \superscript{I}{G}{\delta \boldsymbol{\theta}}{}{t} -\delta \bold{b}_{w,t}\Delta t + \bold{n}_{\boldsymbol{\theta}} \\
        \superscript{I}{G}{\delta \bold{p}}{}{t+1} = \superscript{I}{G}{\delta \bold{p}}{}{t} + \superscript{I}{G}{\delta \bold{v}}{}{t}\Delta t \\
        \superscript{I}{G}{\delta \bold{v}}{}{t+1} = \superscript{I}{G}{\delta \bold{v}}{}{t} + (-\superscript{I}{G}{\bold{R}}{}{t} [\bold{a}_m-\bold{b}_{a,t}]_{\times}\superscript{I}{G}{\delta \boldsymbol{\theta}}{}{t}  -\superscript{I}{G}{\bold{R}}{}{t} \delta \bold{b}_{a,t} )\Delta t \\ + \bold{n}_{v}  \\
        \delta \bold{b}_{a,t+1} = \bold{n}_{b,a} \\
        \delta \bold{b}_{w,t+1} = \bold{n}_{b,w} \\
        \delta \bold{g}_{t+1} = 0 \\
    \end{array}
\label{eq_errors}\end{equation}
where $\bold{n}_{\boldsymbol{\theta}}$, $\bold{n}_v$, $\bold{n}_{b,a}$, and $\bold{n}_{b,w}$ are noise terms for rotation, velocity, accelerometer bias, and gyroscope bias, respectively. The terms with $\delta$ indicate the error values for the corresponding variables, and $\boldsymbol{\theta}_t$ is the rotation vector of $\bold{R}_t$.
The filter setup is similar as in \cite{xu2021fast}. The position, rotation, and velocity states are defined according to the coordinate transformation from IMU to global, with the state vector defined as follows:

\begin{equation}
\renewcommand{\arraystretch}{1.2} \begin{array}{l}
        \bold{x}_t \triangleq 
            \begin{bmatrix}
            \superscript{I}{G}{\bold{R}}{T}{t} ,
            \superscript{I}{G}{\bold{p}}{T}{t} , 
            \superscript{I}{G}{\bold{v}}{T}{t} ,
            \bold{b}_{a,t}^T ,
            \bold{b}_{w,t}^T ,
            \bold{g}_{t}^T
            \end{bmatrix}^T \in \mathcal{M} \\
        \widetilde{\bold{x}}_t \triangleq
            \begin{bmatrix}
                \superscript{I}{G}{\delta \boldsymbol{\theta}}{T}{t}, 
                \superscript{I}{G}{\delta \bold{p}}{T}{t} , 
                \superscript{I}{G}{\delta \bold{v}}{T}{t} , 
                \delta \bold{b}_{a,t}^T , 
                \delta \bold{b}_{w,t}^T , 
                \delta \bold{g}_{t}^T
            \end{bmatrix}^T \in \mathbb{R}^{18}
    \end{array}
\label{eq_imunominal}\end{equation}
where $\bold{x}_t$ is the nominal state and $\widetilde{\bold{x}}_t$ is the error state. The nominal state lies in the manifold ($\mathcal{M}$), while the error states lie in the vector space.

\subsubsection{Image and map point model}
For tightly coupled map matching, an image and map point model must be created. Given the camera intrinsic parameters, $ \superscript{G}{}{\bold{m}}{}{i} $ can be projected onto the camera image. Assuming that the projected feature follows a Gaussian noise distribution, the model is expressed as follows:

\begin{equation}
\superscript{C}{}{\bold{f}}{*}{i} = \pi_1(\superscript{I}{C}{\bold{R}}{}{}  ((\superscript{I}{G}{\bold{R}}{}{t})^T (\superscript{G}{}{\bold{m}}{}{i}- \superscript{I}{G}{\bold{p}}{}{t})) + \superscript{I}{C}{\bold{p}}{}{}) + \bold{n}_f
\end{equation}

\begin{equation}
    \begin{bmatrix}
        u \\
        v
    \end{bmatrix}
     =
    \pi_1(
    \begin{bmatrix}
        X \\
        Y \\
        Z
    \end{bmatrix}
    )
     =
    \begin{bmatrix}
        f_x X/Z + c_x \\
        f_y Y/Z + c_y
    \end{bmatrix} \\[7pt]
\end{equation}
where $f_x$ and $ f_y $ are the focal lengths, and $ c_x $ and $c_y$ are the principal points of the image. $\superscript{I}{C}{\bold{R}}{}{}$ and $\superscript{I}{C}{\bold{p}}{}{}$ are the rotation and translation from the IMU to the camera frame, respectively, and $\bold{n}_f \sim \mathcal{N}(0,\mathbf{R}_f)$. The measurement model, $h_i(\bold{x}_t, \bold{n}_f)$, between predicted 2d feature location and measured 2d feature location is thus represented as follows:

\begin{equation}
\begin{array}{l}
    h_i(\bold{x}_t, \bold{n}_f) \triangleq \superscript{C}{}{\bold{f}}{*}{i} - \superscript{C}{}{\bold{f}}{}{i} \\ \\
      = \pi_1(\superscript{I}{C}{\bold{R}}{}{}  ((\superscript{I}{G}{\bold{R}}{}{t})^T (\superscript{G}{}{\bold{m}}{}{i}- \superscript{I}{G}{\bold{p}}{}{t})) + \superscript{I}{C}{\bold{p}}{}{}) + \bold{n}_f - \superscript{C}{}{\bold{f}}{}{i}
\end{array}
\label{eq_zero_model}
\end{equation}
where $(\superscript{G}{}{\bold{m}}{}{i},  \superscript{C}{}{\bold{f}}{}{i}) \in \mathcal{S}'$.\\[3pt]

\subsubsection{vehicle speed model}
Depending on the vehicle's environment, longitudinal slip may occur, making it risky to use wheel speed alone for measurement. Alternatively, vehicle speed can be used as a measurement. The average of the four or two wheel speeds is assumed to represent the vehicle speed. Representing this as a Gaussian distribution model, the velocity model, $\bold{v}_s$, is expressed as follows:

\begin{equation}
    \bold{v}_s = \superscript{I}{G}{\bold{R}}{}{t}
    \begin{bmatrix}
        v_x \\
        0 \\
        0
    \end{bmatrix}
    + \bold{n}_s \\[5pt]
\end{equation}
where $v_x$ is the average speed of the four or two wheels and $\bold{n}_s \sim \mathcal{N}(0,\mathbf{R}_v)$. Assuming the IMU speed is equal to the vehicle velocity, the measurement model, $h_v(\mathbf{x}_t, \bold{n}_s)$, between velocity model and estimated velocity is represented as follows:

\begin{equation}
    \begin{array}{l}
        h_v(\mathbf{x}_t, \bold{n}_s) \triangleq \bold{v}_s - \superscript{I}{G}{\bold{v}}{}{t} \\ \\ 
        = \superscript{I}{G}{\bold{R}}{}{t} 
        \begin{bmatrix}
            v_x \\
            0 \\
            0
        \end{bmatrix} + \bold{n}_s
        - \superscript{I}{G}{\bold{v}}{}{t} 
    \end{array}
\label{eq_h_v}
\end{equation}

\subsection{Iterative Error State Kalman Filter}
In this study, an Iterative Error State Kalman Filter (IESKF) \cite{xu2022fast} is used, considering the numerous measurement points. First, the propagation phase uses the IMU model, and when measurement points are obtained, the residual computation and update are performed. An iterated update is conducted for optimal state estimation, considering the multiple points.

\subsubsection{Propagation}
Let $\bold{x}_{t-1}$ and $\widehat{\bold{P}}_{t-1}$ be the optimal state and covariance derived from the previous sequence, respectively. The propagation is determined as follows:

\begin{equation}\bar{\bold{x}}_{t} = f(\bold{x}_{t-1},\bold{u}_m).\label{eq_pred1}\end{equation}
\begin{equation}\Bar{\bold{P}}_{t} = \bold{F}_x \widehat{\bold{P}}_{t-1} \bold{F}_x^T + \bold{F}_n \bold{Q}_n \bold{F}_n^T\label{eq_pred2} \\[5pt]
\end{equation}
where $\bar{\bold{x}}_{t}$ and $\bar{\bold{P}}_{t}$ are predicted state and predicted covariance from error model, respectively. Equation \eqref{eq_pred1} represents the state transition process based on the IMU Kinematic model from equation \eqref{eq_imunominal}. Equation \eqref{eq_pred2} represents the covariance for the error state, where $\bold{F}_x$, $\bold{F}_n$, and $\bold{Q}_n$ can be expressed by using equation \eqref{eq_errors}.

\begin{equation} 
\begin{array}{l}
\bold{F}_x = 
    \setlength{\arraycolsep}{3pt}
    \begin{bmatrix}
    \bold{A} & \bold{0}_{3 \times 3} & \bold{0}_{3 \times 3} & \bold{0}_{3 \times 3} & -\bold{I}_{3 \times 3}\Delta t & \bold{0}_{3 \times 3} \\
    \bold{0}_{3 \times 3} & \bold{I}_{3 \times 3} & \bold{I}_{3 \times 3}\Delta t  & \bold{0}_{3 \times 3} & \bold{0}_{3 \times 3} & \bold{0}_{3 \times 3} \\
    \bold{B} & \bold{0}_{3 \times 3} & \bold{I}_{3 \times 3} &  -\superscript{I}{G}{\bold{R}}{}{t}\Delta t  & \bold{0}_{3 \times 3} & \bold{0}_{3 \times 3} \\
    \bold{0}_{3 \times 3} & \bold{0}_{3 \times 3} & \bold{0}_{3 \times 3} & \bold{I}_{3 \times 3} & \bold{0}_{3 \times 3} & \bold{0}_{3 \times 3} \\
    \bold{0}_{3 \times 3} & \bold{0}_{3 \times 3} & \bold{0}_{3 \times 3} & \bold{0}_{3 \times 3} & \bold{I}_{3 \times 3} & \bold{0}_{3 \times 3} \\
    \bold{0}_{3 \times 3} & \bold{0}_{3 \times 3} & \bold{0}_{3 \times 3} & \bold{0}_{3 \times 3} & \bold{0}_{3 \times 3} & \bold{I}_{3 \times 3}
    \end{bmatrix} \\
\\[5pt]
\bold{F}_n = 
    \begin{bmatrix}
    \bold{I}_{3 \times 3} & \bold{0}_{3 \times 3} & \bold{0}_{3 \times 3} & \bold{0}_{3 \times 3} \\
    \bold{0}_{3 \times 3} & \bold{0}_{3 \times 3} & \bold{0}_{3 \times 3} & \bold{0}_{3 \times 3} \\
    \bold{0}_{3 \times 3} & \bold{I}_{3 \times 3} & \bold{0}_{3 \times 3} & \bold{0}_{3 \times 3} \\
    \bold{0}_{3 \times 3} & \bold{0}_{3 \times 3} & \bold{I}_{3 \times 3} & \bold{0}_{3 \times 3} \\
    \bold{0}_{3 \times 3} & \bold{0}_{3 \times 3} & \bold{0}_{3 \times 3} & \bold{I}_{3 \times 3} \\
    \bold{0}_{3 \times 3} & \bold{0}_{3 \times 3} & \bold{0}_{3 \times 3} & \bold{0}_{3 \times 3}
    \end{bmatrix} \\
    \\[5pt]
\bold{Q}_n = 
    \begin{bmatrix}
    \bold{n}_{\boldsymbol{\theta}} \bold{I}_{3 \times 3} & \bold{0}_{3 \times 3} & \bold{0}_{3 \times 3} & \bold{0}_{3 \times 3} \\
    \bold{0}_{3 \times 3} & \bold{n}_v \bold{I}_{3 \times 3} & \bold{0}_{3 \times 3} & \bold{0}_{3 \times 3} \\
    \bold{0}_{3 \times 3} & \bold{0}_{3 \times 3} & \bold{n}_{b,a} \bold{I}_{3 \times 3} & \bold{0}_{3 \times 3} \\
    \bold{0}_{3 \times 3} & \bold{0}_{3 \times 3} & \bold{0}_{3 \times 3} & \bold{n}_{b,w} \bold{I}_{3 \times 3}
    \end{bmatrix} \\ 
\\[5pt]
\text{where }\quad \bold{A}=(\exp[(\bold{w}_m-\bold{b}_{w,t})\Delta t]_\times)^T, \\
\quad \quad \quad \quad \bold{B}=-\superscript{I}{G}{\bold{R}}{}{t}[\bold{a}_m-\bold{b}_{a,t}]_{\times}\Delta t \\    
\end{array}
\\
\label{eq}
\end{equation}

\subsubsection{Residual and Jacobian Computation}
To update the error state, residuals and Jacobians need to be computed. Using the first-order approximation, equation \eqref{eq_zero_model} can be expressed as:

\begin{align}
\begin{split}
        h_i({\bold x}_t,\bold{n}_f)
        & \simeq h_i(\bar{{\bold x}}_t^{\kappa},\bold{0})+ \bold{H}_i^{\kappa}\widetilde{{\bold x}}^{\kappa}_t + \bold{r}_i  \\ 
        & = \bold{z}_i^{\kappa}+\bold{H}_i^{\kappa}\widetilde{{\bold x}}_t^{\kappa} + \bold{r}_i
\end{split}
\label{eq_cam_feat_firstorder}
\end{align}
where $\bar{{\bold x}}_t^{\kappa}$, $\widetilde{{\bold x}}_t^{\kappa}$, $\bold{H}_i^{\kappa}$, and $\bold{r}_i$ are nominal state, error state, Jacobian of the measurement model, and noise vector, respectively. $\bold{z}_i^{\kappa}$ is the residual for the image-map point and expressed as follows using equation \eqref{eq_zero_model}:

\begin{align}
\begin{split}
    \bold{z}_i^{\kappa} &= h_i(\bar{{\bold x}}_t^{\kappa},\bold{0}) \\
    &= \pi_1(\superscript{I}{C}{\bold{R}}{}{}  ((\superscript{I}{G}{\bold{R}}{\kappa}{t})^T (\superscript{G}{}{\bold{m}}{}{i}- \superscript{I}{G}{\bold{p}}{\kappa}{t})) + \superscript{I}{C}{\bold{p}}{}{}) - \superscript{C}{}{\bold{f}}{}{i}
\end{split}
\label{eq_z_i}
\end{align}
Let $\bold{q}_i = [X_i, Y_i, Z_i] = \superscript{C}{I}{\bold{R}}{}{} ((\superscript{G}{I}{\bold{R}}{\kappa}{t})^T (\superscript{G}{}{\bold{m}}{}{i} - \superscript{G}{I}{\bold{p}}{\kappa}{t})) + \superscript{C}{I}{\bold{p}}{}{}$. Then, the Jacobian is expanded as follows:
 
\begin{align}
\begin{split}
    \bold{H}_i^\kappa &= {\displaystyle \frac{\partial h_i(\bold{x},\bold{n}_f)}{\partial \delta \bold{x}}}\Bigr| 
     _{\bold{x}=\Bar{\bold{x}}_t^\kappa} \\ &=
    \left[\begin{array}{cccccc}
       {\displaystyle \frac{\partial  h_i}{\partial \delta \boldsymbol{\theta}_t}} & { \displaystyle \frac{ \partial h_i}{\partial \delta \bold{p}_t}} & \bold0_{2 \times 3} & \bold0_{2 \times 3} & \bold0_{2 \times 3} & \bold0_{2 \times3} 
\end{array}\right]
\end{split}
\label{eq_hi_0}
\end{align}

\begin{flalign}
\begin{split}
     \text{where}\quad &\frac{\partial  h_i}{\partial \delta \bold{p}_t} = {\partial  h_i(\bold{x},\bold{n}_f) \over \partial \bold{q}_i} {\partial \bold{q}_i \over \partial \delta \bold{p}_t} \\
     & \quad \quad \Hquad= -{\displaystyle \frac{\partial  h_i(\bold{x},\bold{n}_f)}{ \partial \bold{q}_i}} \superscript{I}{G}{\bold{R}}{}{} (\superscript{I}{G}{\bold{R}}{\kappa}{t} )^T
    \\[7pt]
     &\frac{\partial  h_i }{ \partial \delta \boldsymbol{\theta}_t} = {\partial  h_i(\bold{x},\bold{n}_f) \over \partial \bold{q}_i} {\partial \bold{q}_i \over \partial \delta \boldsymbol{\theta}_t} \\ 
     & \quad \quad \Hquad= \displaystyle{ {\partial  h_i(\bold{x},\bold{n}_f) \over \partial \bold{q}_i} }\superscript{I}{C}{\bold{R}}{}{} [(\superscript{I}{G}{\bold{R}}{\kappa}{t} )^T (\superscript{G}{}{\bold{m}}{}{i} - \superscript{I}{G}{\bold{p}}{\kappa}{t} )]_\times
    \\[7pt]
    &{\partial  h_i(\bold{x},\bold{n}_f) \over \partial \bold{q}_i} =
    \begin{bmatrix}
        \begin{array}{ccc}
            f_x/Z_i & 0 & -f_xX_i/(Z_i)^2 \\
            0 & f_y/Z_i & -f_yY_i/(Z_i)^2 
        \end{array}
    \end{bmatrix}
\end{split} &&
\end{flalign}

Similarly, for equation \eqref{eq_h_v}, first-order approximation is applied:

\begin{align}
\begin{split}
        h_v({\bold x}_t,\bold{n}_s)
        &\simeq h_v(\bar{{\bold x}}_t^{\kappa},0)+ \bold{H}_v^{\kappa}\widetilde{{\bold x}}^{\kappa}_t + \bold{r}_v  \\ 
        &= \bold{z}_v^{\kappa}+\bold{H}_v^{\kappa}\widetilde{{\bold x}}_t^{\kappa} + \bold{r}_v
\end{split}
\label{eq_hv_0}
\end{align}
where $\bold{H}_v^{\kappa}$ and $\bold{r}_v$ are Jacobian from equation \eqref{eq_h_v} and noise vector of velocity model, respectively. $\bold{z}_v^{\kappa}$ is the residual for velocity and expressed as:

\begin{align}
\begin{split}
    \bold{z}_v^{\kappa} &= h_v(\bar{{\bold x}}_t^{\kappa},\bold{0}) \\
    &= \superscript{I}{G}{\bold{R}}{}{t} 
        \begin{bmatrix}
            v_x \\
            0 \\
            0
        \end{bmatrix}
        - \superscript{I}{G}{\bold{v}}{}{t} 
    \\[7pt]
\end{split}
\label{eq_z_v}
\end{align}

The Jacobian $\bold{H}_v^\kappa$ is expressed as:

\begin{align}
\begin{split}
    \bold{H}_v^\kappa &= {\displaystyle \frac{\partial h_v(\bold{x},\bold{n}_s)}{\partial \delta \bold{x}}}\Bigr| 
     _{\bold{x}=\Bar{\bold{x}}_t^\kappa} \\ &=
    \left[\begin{array}{cccccc}
       {\displaystyle \frac{\partial  h_v}{\partial \delta \boldsymbol{\theta}_t}} & \bold0_{2,3} & { \displaystyle \frac{ \partial h_i}{\partial \delta \bold{v}_t}} & \bold0_{2 \times 3} & \bold0_{2 \times 3} & \bold0_{2 \times 3} 
\end{array}\right]
\end{split}
\label{eq_hv_0}
\end{align}

\begin{flalign}
\begin{split}
    \text{where} \quad \frac{\partial  h_v}{\partial \delta \bold{v}_t} &= -\bold{I}_3
    \\
     \frac{\partial  h_v}{\partial \delta \boldsymbol{\theta}_t} &= -\superscript{I}{G}{\bold{R}}{\kappa}{t}[[v_s \quad 0 \quad 0]^T]_\times \\[7pt]
\end{split} &&
\end{flalign}

\subsubsection{Iterated update}
The propagated state, $\bar{\bold{x}}_{t}$, and covariance, $\bar{\bold{P}}_{t}$, follow a prior Gaussian distribution concerning the unknown state $\bold{x}_{t}$, and are expressed as follows \cite{xu2022fast}:

\begin{align}
\begin{split}
    \bold{x}_t \boxminus \bar{\bold{x}}_t &= (\bar{\bold{x}}_t^\kappa \boxplus \widetilde{\bold{x}}_t^\kappa) \boxminus \bar{\bold{x}}_t
    = \bar{\bold{x}}_t^\kappa \boxminus \bar{\bold{x}}_t + \bold{J}^\kappa \widetilde{\bold{x}}_t^\kappa \\
    & \sim \mathcal{N}(0,\bar{\bold{P}}_k) \\[1pt]
\label{eq_distribution_prior}
\end{split} 
\end{align}
where $\boxplus$ and $\boxminus$ are box-plus and box-minus operator in manifold, and $\bold{J}^\kappa$ is the partial differentiation of $(\bar{\bold{x}}_t^\kappa \boxplus \widetilde{\bold{x}}_t^\kappa) \boxminus \bar{\bold{x}}_t$ with respect to $\widetilde{\bold{x}}_t^\kappa$ evaluated at zero

\begin{align}
\begin{split}
\bold{J}^\kappa = 
    \begin{bmatrix}
    \bold{A}( \superscript{I}{G}{\delta \boldsymbol{\theta}}{\kappa}{t} )^{-T} & \bold{0}_{3 \times 15}\\
    \bold{0}_{15 \times 3} & \bold{I}_{15 \times 15} 
    \end{bmatrix}
\end{split} 
\end{align}
\begin{algorithm}[H]
\caption{Localization process}\label{alg:cap}
\begin{algorithmic}[1]
    \Statex \textbf{Input:} Last output $\widehat{\bold{x}}_{t-1}$ and $\widehat{\bold{P}}_{t-1}$;
    \Statex \quad \quad \quad Current image $\superscript{C}{C}{\bold{I}}{}{t}$;
    \Statex \quad \quad \quad Current IMU data $(\bold{a}_m,\bold{w}_m)$;
    \Statex \quad \quad \quad Topological map $\mathcal{T}$
    \STATE $\bar{\bold{x}}_{t}$, $\bar{\bold{P}}_{t} \gets$ Forward propagation via \eqref{eq_pred1} and \eqref{eq_pred2}
    \STATE $\superscript{C^*}{}{\mathcal{N}}{}{k} \gets$ get node from $\mathcal{T}$ by kd-tree search with $\bar{\bold{x}}_{t}$
    \STATE $ \mathcal{F}_C,  \mathcal{F}_{C^*} \gets$ correspondence matching from $ \superscript{C}{C}{\bold{I}}{}{t},  \superscript{C^*}{C}{\bold{I}}{}{k}$
    \STATE $ \mathcal{S} \gets $ outlier removal via \eqref{eq_F_C_start_dash},\eqref{eq_mean_pixel} and \eqref{eq_matching_set_2d_2d}
    \STATE $ \mathcal{S'} \gets $ 3d map point restoration from $ \mathcal{S}  $, $\superscript{C^*}{G}{\bold{T}}{}{k}$, and $ \superscript{C^*}{}{\bold{D}}{}{k} $
    \STATE $\kappa \gets -1$; $ \bar{\bold{x}}_{t}^{\kappa = 0} \gets \bar{\bold{x}}_{t} $
    \REPEAT
    \STATE $\kappa \gets \kappa + 1$
    \STATE $\bold{z}_i^\kappa, \bold{H}_i^\kappa \gets$ get residual and Jacobian via \eqref{eq_z_i} and \eqref{eq_hi_0} for matching points $\mathcal{S}'$
    \STATE $\bold{z}_v^\kappa, \bold{H}_v^\kappa \gets$ get residual and Jacobian via \eqref{eq_z_v} and \eqref{eq_hv_0}
    \STATE $\bold{z,H,R} \gets $ concatenate via \eqref{eq_HRz}
    \STATE $\bold{P} \gets (\bold{J}^\kappa)^{-1} \Bar{\bold{P}}_t (\bold{J}^\kappa)^{-T}$
    \STATE $\bar{\bold{x}}_{t}^{\kappa+1} \gets $ state update via \eqref{eq_k}, \eqref{eq_x_tilde} and \eqref{eq_x_bar1}
    \UNTIL $||\bar{\bold{x}}_{t}^{\kappa+1} \boxminus \bar{\bold{x}}_{t}^{\kappa}|| < \epsilon$
    \STATE $\widehat{\bold{x}}_{t} \gets \bar{\bold{x}}_{t}^{\kappa+1} $; $ \widehat{\bold{P}}_{t} \gets \bold{(I-KH)P}$
    \Statex \textbf{Output:}  $\widehat{\bold{x}}_{t}$ and $\widehat{\bold{P}}_{t}$
\end{algorithmic}
\end{algorithm}
where $A(\cdot)^{-1}$ is defined in \cite{he2021kalman}, and $\superscript{I}{G}{\delta \boldsymbol{\theta}}{\kappa}{t} = \superscript{I}{G}{\bar{\bold{R}}}{\kappa}{t} \boxminus \superscript{I}{G}{\bar{\bold{R}}}{}{t}$ is the error state of the IMU's rotation in iteration $\kappa$.
The state distribution from measurements can be derived as follows.

\begin{align}
\begin{split}
        - \bold{r}_i& = \bold{z}_i^{\kappa}+\bold{H}_i^{\kappa}\widetilde{{\bold x}}_k^{\kappa} \sim \mathcal{N}(0,\bold{R}_i) \\
        - \bold{r}_v& = \bold{z}_v^{\kappa}+\bold{H}_v^{\kappa}\widetilde{{\bold x}}_k^{\kappa} \sim \mathcal{N}(0,\bold{R}_v) \\[7pt]
\end{split}
\label{eq_distribution_measurement}
\end{align}

Using equations \eqref{eq_distribution_prior} and \eqref{eq_distribution_measurement}, the posteriori distribution of state $\bold{x}_t$ can be determined. The Maximum A Posteriori (MAP) estimate is given by:

\begin{align}
\begin{split}
    \min\limits_{\widetilde{\bold x}_k^{\kappa}}(||\bold x_k \boxminus \Bar{\bold x}_k||^2_{\Bar{\bold{P}}_k^{-1}} 
    &+ \Sigma^N_{i=1}|| \bold{z}_i^{\kappa}+\bold{H}_i^{\kappa}\widetilde{{\bold x}}_k^{\kappa} ||^2_{\bold{R}_i^{-1}} \\
    &+ || \bold{z}_v^{\kappa}+\bold{H}_v^{\kappa}\widetilde{{\bold x}}_k^{\kappa} ||^2_{\bold{R}_v^{-1}})\\[7pt]
\end{split}
\label{eq_map}
\end{align}
where $||\bold x||^2_{\textbf{Q}} = \bold x^T \textbf{Q} \bold x$. This MAP problem can be solved using the iterated Kalman filter and is expressed below:

\begin{equation}
    \widetilde{\bold{x}}_{t}^{\kappa} = -\bold{Kz} - \bold{(I-KH)} (\bold{J^\kappa})^{-1}(\bar{\bold{x}}_{t}^{\kappa} \boxminus \bar{\bold{x}}_{t})\label{eq_x_tilde}
\end{equation}
\begin{equation}
    \bold{K} = (\bold{H}^T \bold{R}^{-1} \bold{H} + \bold{P}^{-1})^{-1} \bold{H}^T \bold{R}^{-1} \label{eq_k}
\end{equation}
\begin{align}
\begin{split}
\bold{P} &= (\bold{J}^\kappa)^{-1} \Bar{\bold{P}}_t (\bold{J}^\kappa)^{-T}
\end{split}
\label{eq_P_}
\end{align}
\begin{equation}
    \text{where} \quad \bar{\bold{x}}_{t}^{\kappa+1} = \bar{\bold{x}}_{t}^{\kappa} \boxplus \widetilde{\bold{x}}_{t}^{\kappa} \qquad \qquad \qquad \qquad \qquad \qquad \Hquad \label{eq_x_bar1}
\end{equation}
\begin{align}
\begin{split}
\bold{z} &= [(\bold{z}_1^\kappa)^T,\cdots,(\bold{z}_N^\kappa)^T, (\bold{z}_v^\kappa)^T]^T, \\
\bold{H} &= [(\bold{H}_1^\kappa)^T,\cdots,(\bold{H}_N^\kappa)^T, (\bold{H}_v^\kappa)^T]^T, \\
\bold{R} &= [(\bold{R}_1)^T,\cdots,(\bold{R}_N)^T, (\bold{R}_v)^T]^T, \\[5pt]
\end{split}
\label{eq_HRz}
\end{align}

Here, $\bold{K}$ is the Kalman gain. To speed up the calculation with a high number of matching points, equation \eqref{eq_k} is modified as per \cite{xu2021fast}.
The above correction process repeats until the convergence ($||\bar{\bold{x}}_{t}^{\kappa + 1} \boxminus \bar{\bold{x}}_{t}^{\kappa}|| < \epsilon$). Upon convergence, the optimal state and covariance are expressed as follows:

\begin{equation}
    \widehat{\bold{x}}_{t} = \bar{\bold{x}}_{t}^{\kappa+1};
    \widehat{\bold{P}}_{t} = \bold{(I-KH)P}\label{eq_result_state} \\[3pt]
\end{equation}

The overall localization process is summarized in Algorithm 2.

\section{Experimental results}
The proposed algorithm is validated using the Complex Urban Dataset \cite{jjeong-2019-ijrr} and our collected data in experimentally challenging scenario. Absolute Pose Error (APE) is used as a evaluation metric to verify the localization algorithm or the topological map against the ground truth. Furthermore, the evaluation metric is divided into rotation (APEr) and translation (APEt) to assess the performance.

\begin{align}
\begin{split}
\mathrm{APEt} &= \frac{1}{K} \Sigma^K_{t=0}||\superscript{I}{G}{\bold{p}}{}{t} - \superscript{I}{G}{\bold{p}}{}{t, true}||_2 \\ 
\mathrm{APEr} &= \frac{1}{K} \Sigma^K_{t=0}(\arccos(0.5*\mathrm{trace}(\superscript{I}{G}{\bold{R}}{T}{t,true} \superscript{I}{G}{\bold{R}}{}{t})-1)) 
\end{split}
\label{eq_P_}
\end{align}
where $\superscript{I}{G}{\bold{p}}{}{t, true}$ and $\superscript{I}{G}{\bold{R}}{}{t,true}$ are true position and true rotatation of IMU frame in global frame, respectively. The proposed algorithm is validated using an Intel i7-6700K 4.00GHz CPU with 8 cores, and an Nvidia Geforce RTX 4080 Super GPU. 

\subsection{Evaluation on complex urban dataset}
The Complex Urban Dataset is well-suited for validating localization algorithms as it provides both LiDAR point cloud maps and ground truth pose trajectories. Additionally, by comparing the ground truth path with the poses in the topological map, it is possible to validate the topological map proposed in this study. Two specific scenarios are evaluated, where the scenario locations are identical but the logging times differ, allowing for the assessment of relocalization capabilities.
\subsubsection{Topological map validation}
\begin{figure}[!t]
\centering
\includegraphics[width=3.9in]{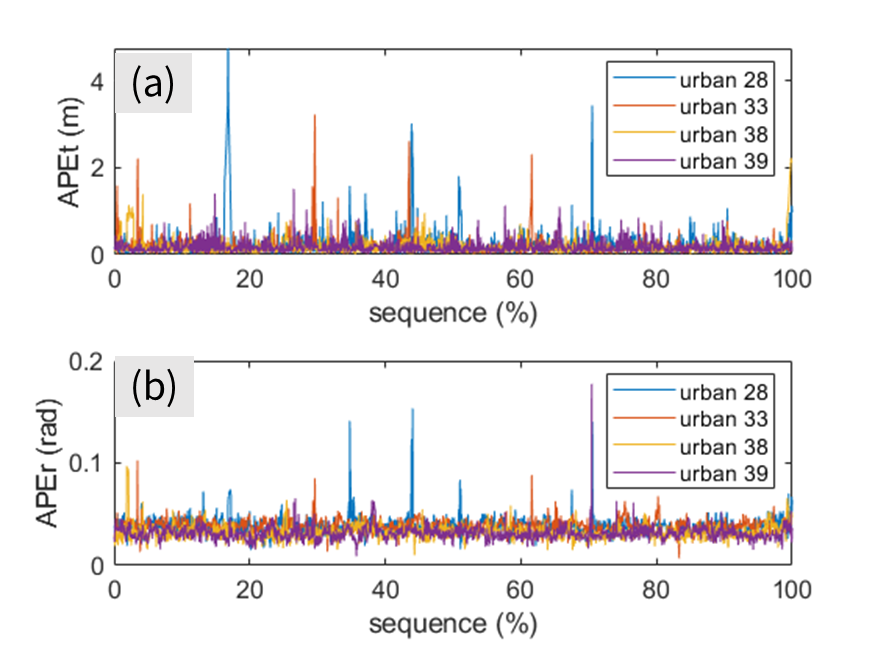}
\caption{Topological map pose evaluation with ground truth pose.}
\label{top_map_res}
\end{figure}

Figure \ref{top_map_res} presents the comparison between the camera poses generated during the creation of the topological map and the ground truth poses. The ground truth poses are transformed using the vehicle-to-camera calibration data provided in the Complex Urban Dataset. In Figure \ref{top_map_res}-(a), most sequences show values close to zero, indicating good agreement. However, some local sequences exhibit differences, suggesting errors occurred during the creation of the topological map. Nodes with large errors increase the likelihood of localization errors. Thus, this experiment excludes sections with significant pose discrepancies.

Figure \ref{top_map_res}-(b) shows a bias of approximately 0.03 to 0.04 radians across all scenarios. Upon analysis, this issue is attributed to errors in the vehicle-to-camera extrinsic parameters within the Complex Urban Dataset. If the rotation value of the ground truth were inaccurate, it would exhibit noise rather than a consistent bias. However, in the scenarios validated in this study, all results indicated the presence of bias with minimal noise. This implies that while the rotation values of the ground truth are accurate, there is a bias in the rotation. The ground truth pose is based on the rear wheels of the vehicle, and if this rotation has a bias, it can be corrected quickly with the vehicle's movement. Therefore, it is concluded that the rotation bias in the camera pose, which is a combination of the ground truth pose and extrinsic calibration, is likely due to issues with the rotation in the extrinsic calibration. This can be identified by comparing sequences where the topological map is well-constructed.

\begin{figure}[!t]
\centering
\includegraphics[width=3.5in]{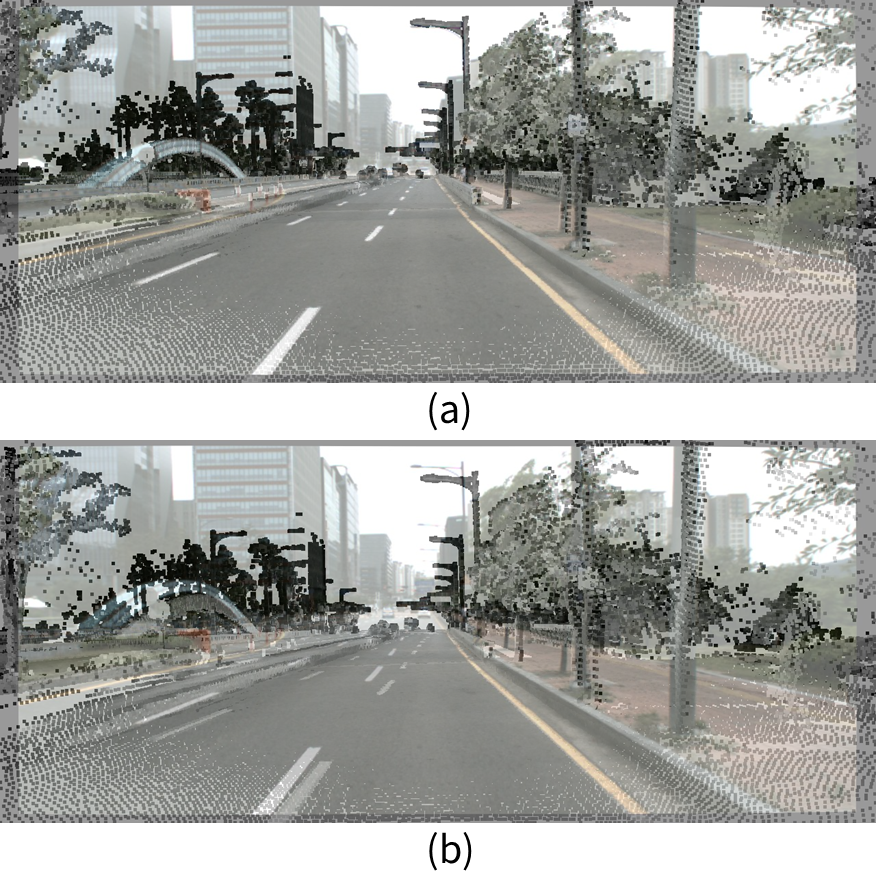}
\caption{Overlapped camera image and corresponding point cloud intensity image in (a) calculated pose and (b) ground truth pose.}
\label{compare_top_ref}
\end{figure}

Figure \ref{compare_top_ref} overlays camera images with point cloud intensity images derived from both the estimated camera poses and the ground truth poses, respectively, in sequences with low APEt values. Figure \ref{compare_top_ref}-(a) shows that the roadmarks captured in the camera image align well with those in the point cloud intensity image. In contrast, Figure \ref{compare_top_ref}-(b) shows slight discrepancies. Comparing these results with Figure \ref{top_map_res}-(b) suggests that the issue likely lies with the rotation of the ground truth poses, leading to a constant bias in the rotation due to errors in the vehicle-to-camera extrinsic parameters.

\subsubsection{Self-Localization Results}
Table \ref{table_self_loc_perf} presents the results of self-localization where the logging data used for map creation is the same as the logging data used for the localization test. All scenarios pertain to urban roads, with results shown for each case. OpenVINS \cite{geneva2020openvins} results are obtained by fixing the initial pose values as a reference in visual SLAM. The "low-cost GPS" refers to GPS data used in commercial vehicles, while "VRS-GPS" denotes high-precision GPS. The case of "OpenVINS (w ICP)" involves accumulating the feature map output by the OpenVINS algorithm and performing ICP map matching with a LiDAR prior map.

Due to the odometry characteristics, OpenVINS exhibits significant errors resulting from cumulative errors. For GPS, VRS-GPS shows more accurate positioning than commercial GPS, but urban road scenarios still introduce errors due to the multipath effect. While "OpenVINS (w ICP)" shows improved performance through global localization, errors can still be significant due to incorrect matching or computational delays caused by ICP. In contrast, the algorithm proposed in this study demonstrates superior performance across all scenarios.

The proposed algorithm enhances performance by incorporating vehicle speed. The performance without considering vehicle speed is shown in "proposed (w/o speed)." It is observed that removing speed results in decreased performance compared to when it is included. In some scenarios, the performance without speed is even worse than that of VRS-GPS. The reduction in performance when excluding speed might be due to the lack of full rank observability. Empirically, this study finds that degeneracy issues tend to arise when the number of features is insufficient or when the correspondence matching occurs only in certain regions of the image. Addressing this issue will be left for future research.

\begin{table*}[t!]
\centering
\caption{Localization Performance in Complex Urban Dataset}
\label{table_self_loc_perf}
\setlength{\tabcolsep}{10pt}
\begin{tabular}{c|c|c|c|c|c|c|c}
\hline
 Sequence                   &           & OpenVins & Low cost GPS & VRS-GPS &OpenVins (w ICP)&  Proposed (w/o speed)&  Proposed \\ \hline
\multirow{2}{*}{28 (11.47km)}& APEt (m) & 34.63     & 4.08        & 2.53    & 5.12           & 1.27                 & \textbf{0.818} \\
                            & APEr (rad)& 0.0544    & 0.290       & -       & 0.0686         & 0.0537               & \textbf{0.0455} \\ \hline
\multirow{2}{*}{33 (7.6km)} & APEt (m)  & over 100  & 4.18        & 1.60    & 17.46          & 2.18                 & \textbf{0.827} \\
                            & APEr (rad)& 1.57      & 0.297       & -       & 0.0789         & 0.0473               & \textbf{0.0388} \\ \hline
\multirow{2}{*}{38 (11.42km)} & APEt (m)& 63.97     & 4.00        & 2.95    & 4.85           & 1.45                 & \textbf{0.667} \\
                            & APEr (rad)& 0.119     & 0.291       & -       & 0.0708         & 0.0557               & \textbf{0.0408} \\ \hline
\multirow{2}{*}{39 (11.06km)} & APEt (m)& 22.83     & 4.06        & 2.13    & 4.47           & 1.16                 & \textbf{0.625} \\
                            & APEr (rad)& 0.0453    & 0.342       & -       & 0.0657         & 0.0435               & \textbf{0.0368} \\ 
\end{tabular}
\end{table*}

\subsubsection{Relocalization Analysis}
\begin{figure*}[!t]
\centering
\includegraphics[width=7.0in]{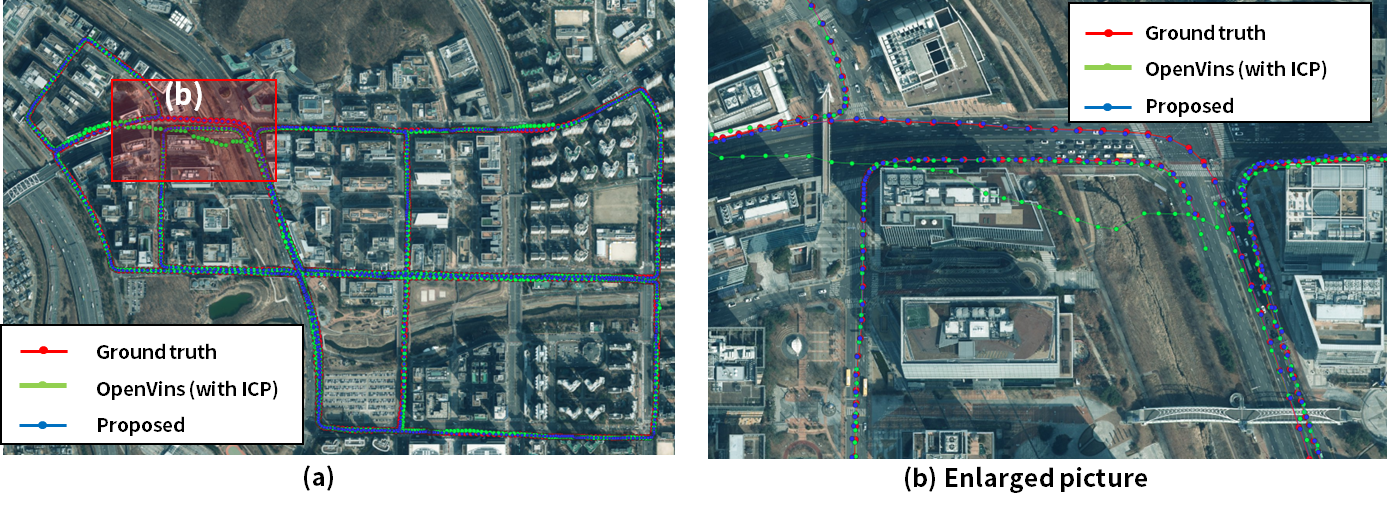}\caption{Relocalization results from sequence urban38 utilizing map urban28 in (a) total sequence and (b) enlarged with ICP fail case }
\label{relocalization_results}
\end{figure*}

In the Complex Urban Dataset, scenarios 28 and 38 have nearly identical routes despite being collected at different times, resulting in similarly constructed maps. This study uses these two scenarios for relocalization analysis. Figure \ref{relocalization_results} shows the position results for the urban38 scenario using the map from urban28.

In Figure \ref{relocalization_results}-(a), the proposed algorithm consistently yields accurate localization results, whereas OpenVINS (w ICP) shows incorrect results in some sections due to improper ICP matching. Figure \ref{relocalization_results}-(b) zooms in on a specific area in (a), illustrating instances where ICP matching fails, leading to incorrect road estimations or positions outside the road. In contrast, the proposed algorithm closely follows the ground truth.

Table \ref{table_relocalize} presents the quantitative evaluation results of relocalization. Although the Point Cloud map for OpenVINS (w ICP) was created at a different time, the results are similar to the self-localization described in the previous section. The proposed algorithm shows slightly decreased performance compared to self-localization but still outperforms OpenVINS (w ICP).

\begin{table}[t!]
\centering
\caption{Relocalization Performance in Complex Urban Dataset}
\label{table_relocalize}
\setlength{\tabcolsep}{10pt}
\begin{tabular}{c|c|c|c|c}
\hline
 Map &Seq.&    &OpenVins (w ICP)&  Proposed \\ \hline
\multirow{4}{*}{28} & \multirow{4}{*}{38} & APEt (m) & 4.35 & 1.91 \\
                    &                     & lon. (m) & 3.43 & 1.34 \\ 
                    &                     & lat. (m) & 1.67 & 1.11 \\
                    &                     & APEr (rad) & 0.0649 & 0.0463 \\ \hline
\multirow{4}{*}{38} & \multirow{4}{*}{28} & APEt (m) & 3.83 & 1.88 \\
                    &                     & lon. (m) & 3.26 & 1.22 \\
                    &                     & lat. (m) & 1.30 & 1.07 \\
                    &                     & APEr (rad) & 0.0597 & 0.0464 \\ \hline
\end{tabular}
\end{table}

\subsection{Verification from experimentally challenging scenario}
The experimentally challenging scenario involves a highway with tunnels. This scenario includes both a LiDAR prior map and an HD map, which will be used to validate the localization. Due to the nature of tunnels, GPS reception is difficult and it is challenging to generate the ground truth. This paper demonstrates that not only a topological map can be created through topological map generation, but also ground truth can be also provided. Moreover, the localization results on an actual highway will be presented using the generated topological map. In the collected data, the test vehicle setup was equipped with the GPS, the front view camera, and the IMU units. The equipped GPS and IMU can exchange information with each other to receive a dead-reckoning solution. To enhance the positional accuracy of the GPS, Network RTK was employed. The chassis CAN data was also collected to measure vehicle speed signals.
\subsubsection{Topological map qualitative evaluation}
\begin{figure*}[!t]
\centering
\includegraphics[width=7.0in]{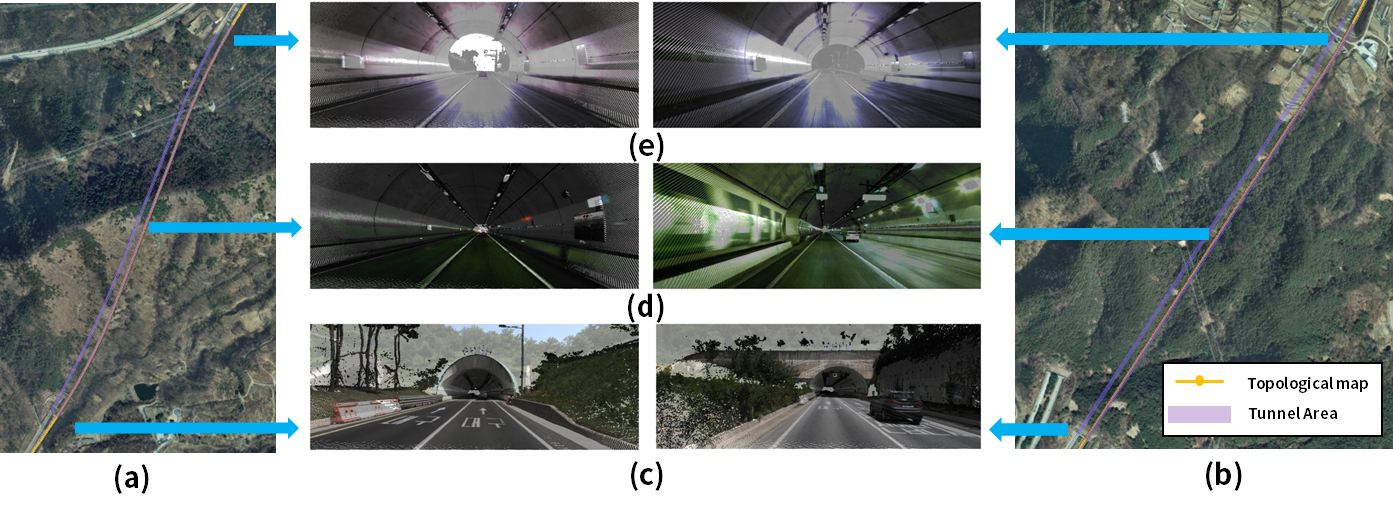}\hfil
\caption{Topological map validation from overlapping camera image and corresponding point cloud intensity image in (a) curved tunnel and (b) straight tunnel. The results is shown at the (c) entrance, (d) intermediate, and (e) exit of the tunnels.}
\label{fig_topol_map_tech}
\end{figure*}

Fig. \ref{fig_topol_map_tech} presents the validation results for topological map generation. Figs. \ref{fig_topol_map_tech}-(a) and (b) depict a curved tunnel and a straight tunnel, respectively. The orange colored lines with dot represent the positions in the topological map, showing that poses are generated from the entrance to the exit. For each case, the entrance is shown in Fig. \ref{fig_topol_map_tech}-(c), the intermediate section in Fig. \ref{fig_topol_map_tech}-(d), and the exit in Fig. \ref{fig_topol_map_tech}-(e), with camera images and LiDAR intensity images overlapped. In all cases, looking at the walls and lane markings, the camera image and LiDAR intensity image align closely, indicating that the camera pose was accurately determined. Notably, Fig. \ref{fig_topol_map_tech}-(e) also shows successful results, suggesting that a topological map can be generated using initial position and odometry without GPS. This result implies that topological map generation can effectively find poses even in tunnels where obtaining ground truth poses is challenging.

\subsubsection{Localization evaluation}
\begin{figure*}[!t]
\centering
\includegraphics[width=7.0in]{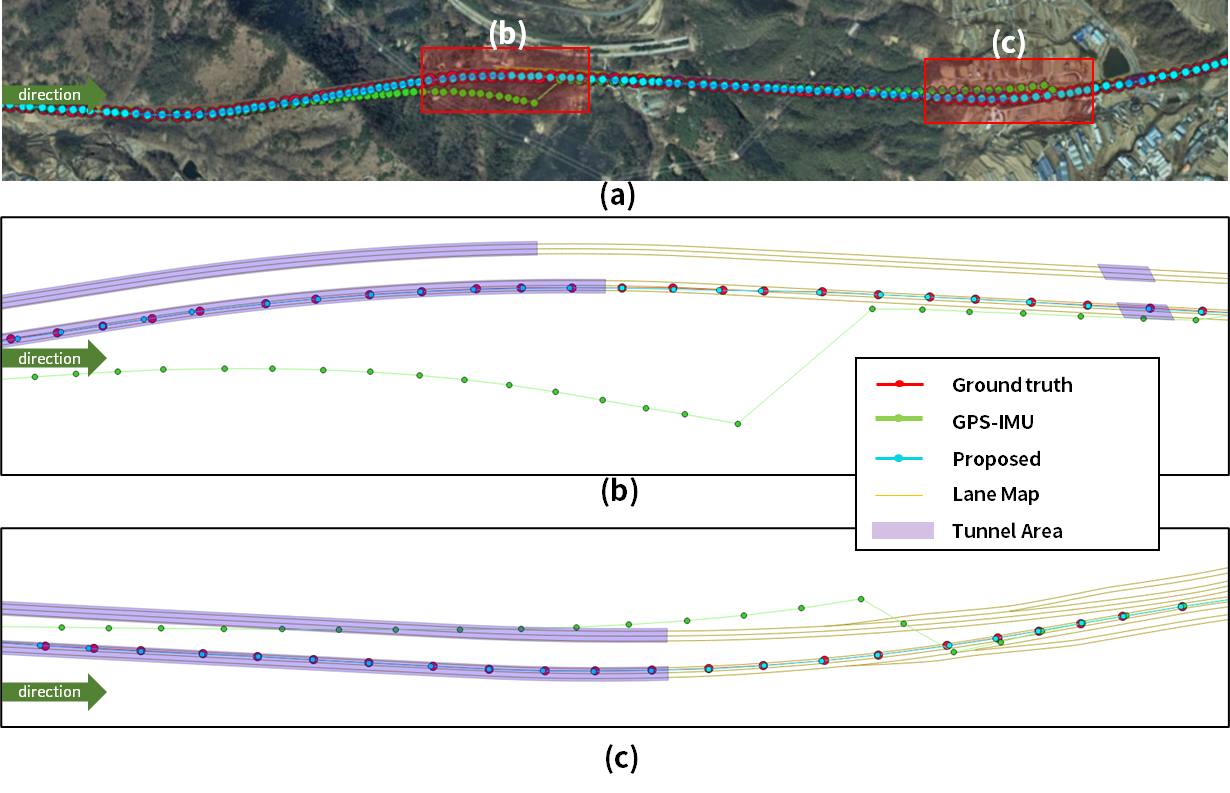}\hfil
\caption{Localization results in tunnel area in (a) total map, (b) curved tunnel, and (c) straight tunnel}
\label{fig_localization_tech}
\end{figure*}

Fig. \ref{fig_localization_tech} shows the position results for a scenario involving two tunnel sections. The ground truth position is based on the previously described topological map, while the GPS-IMU position is derived from integrating RTK-GPS and dead-reckoning solutions. Fig. \ref{fig_localization_tech}-(a) displays the results from left to right, and Figs. \ref{fig_localization_tech}-(b) and (c) provide zoomed-in views of the tunnel exits. At the tunnel exits, the GPS-IMU solution struggles with position estimation despite using dead-reckoning. However, the proposed algorithm accurately localizes within the lane. The evaluation results for the entire section are summarized in Table \ref{table_localize_tech}. Compared to the GPS-IMU, the proposed algorithm demonstrates superior performance. Notably, the proposed algorithm effectively reduces the longitudinal error in localization, a critical issue in tunnels.

\begin{table}[t!]
\centering
\caption{Localization Performance evaluation in tunnel area}
\label{table_localize_tech}
\setlength{\tabcolsep}{10pt}
\begin{tabular}{c|c|c}
\hline
          &GPS-IMU&  Proposed \\ \hline
 APEt (m) & 17.63  & \textbf{1.58} \\
lon. (m) & 10.16  & \textbf{1.45} \\ 
lat. (m) & 13.67  & \textbf{0.293} \\
 APEr (rad) & 0.0682 & \textbf{0.0246} \\ \hline

\end{tabular}
\end{table}

\section{Conclusion}

This paper presented a tightly-coupled, speed-aided monocular visual-inertial localization algorithm utilizing a topological map structure. The proposed method effectively transforms LiDAR point cloud maps into a topological format, allowing for efficient map matching and robust pose estimation. The proposed approach addresses the challenges associated with using high-cost sensors by incorporating relatively inexpensive camera-based localization, enhanced with vehicle speed measurements. Through experiments on the Complex Urban Dataset, it is demonstrated that proposed algorithm outperforms traditional methods like OpenVINS with ICP in both self-localization and relocalization scenarios. In experiments on our collected data, it is also demonstrated that the localization can be accurately performed even if in the challenging scenario, such as tunnel or instability of GPS reception. Both results indicate that the proposed method provides accurate localization suitable for autonomous driving applications.

\bibliographystyle{IEEEtran}
\bibliography{ref}

\end{document}